%% file: SBP-YOLO-arxiv.tex
\documentclass{article}

\usepackage{arxiv}

\usepackage[utf8]{inputenc} % allow utf-8 input
\usepackage[T1]{fontenc}    % use 8-bit T1 fonts
\usepackage{hyperref}       % hyperlinks
\usepackage{url}            % simple URL typesetting
\usepackage{booktabs}       % professional-quality tables
\usepackage{amsfonts}       % blackboard math symbols
\usepackage{nicefrac}       % compact symbols for 1/2, etc.
\usepackage{microtype}      % microtypography
\usepackage{lipsum}
\usepackage{graphicx}
\usepackage{latexsym}
\usepackage[misc]{ifsym}% 添加脚注预定义设置，换行以后可实现格式对齐
\usepackage{amssymb}
\usepackage{cite} % 导入引用的包，能够使用\cite
\usepackage{amsmath}
\usepackage{caption}  % 确保导言区引入该宏包
\usepackage[hang]{footmisc}% 添加脚注对齐设置
\usepackage{float}  % 导言区添加
\usepackage{hyperref}  % 支持 \url 和超链接
\usepackage{abstract}
\graphicspath{ {./images/} }

\title{
SBP-YOLO:A Lightweight Real-Time Model for Detecting Speed Bumps and Potholes toward Intelligent Vehicle Suspension Systems
}

\author{
Chuanqi Liang\textsuperscript{1} \quad
Jie Fu\textsuperscript{1} \quad
Miao Yu\textsuperscript{1}\thanks{Corresponding author: yumiao@cqu.edu.cn} \quad
Lei Luo\textsuperscript{1} \\
\textsuperscript{1}Key Laboratory for Optoelectronic Technology and Systems, Ministry of Education,\\
College of Optoelectronic Engineering, Chongqing University, Chongqing, China \\
\texttt{20230801037@stu.cqu.edu.cn}, \texttt{fujie@cqu.edu.cn}, \texttt{llei@cqu.edu.cn}, \texttt{yumiao@cqu.edu.cn} 
}

\begin{document}
\maketitle
\begin{abstract}
Speed bumps and potholes are the most common road anomalies, significantly affecting ride comfort and vehicle stability. 
Preview-based suspension control mitigates their impact by detecting such irregularities in advance and adjusting suspension parameters proactively. 
Accurate and real-time detection is essential, 
but embedded deployment is constrained by limited computational resources and the small size of targets in input images.
To address these challenges, this paper proposes SBP-YOLO, an efficient detection framework for speed bumps and potholes in embedded systems. 
Built upon YOLOv11n, it integrates GhostConv and VoVGSCSPC modules in the backbone and neck to reduce computation while enhancing multi-scale semantic features. 
A P2-level branch improves small-object detection, and a lightweight and efficient detection head (LEDH) maintains accuracy with minimal overhead. 
A hybrid training strategy further enhances robustness under varying road and environmental conditions, combining NWD loss, BCKD knowledge distillation, and Albumentations-based augmentation. 
Experiments show that SBP-YOLO achieves 87.0\% mAP, outperforming the YOLOv11n baseline by 5.8\%. 
After TensorRT FP16 quantization, it runs at 139.5 FPS on Jetson AGX Xavier, yielding a 12.4\% speedup over the P2-enhanced YOLOv11. 
These results demonstrate the framework's suitability for fast, low-latency road condition perception in embedded suspension control systems.      
\end{abstract}

% keywords can be removed

\section{Introduction}
Adjustable suspension systems, such as magnetorheological semi-active~\cite{li2025robust,yu2025research} 
and fully active suspensions~\cite{yu2024advances}, enhance ride comfort and handling stability by enabling real-time damping control, 
while fully active systems can additionally adjust stiffness to adapt to varying road conditions~\cite{ferhath2024evolution}. 
Conventional control strategies, typically based on suspension deflection and sprung-mass acceleration feedback, 
improve vibration suppression to some extent. 
However, without road preview information, these approaches remain reactive, 
limiting the ability to anticipate disturbances and thereby constraining the full potential of advanced suspension control~\cite{achnib2023discrete}. 
Among various road disturbances, speed bumps and potholes represent common yet critical anomalies that directly affect ride quality~\cite{wambold2009roughness}. 
Early detection of such features enables predictive suspension control, 
allowing damping strategies to be adjusted in advance~\cite{li2024explicit,jung2025preview,wang2025road}. 
Existing methods for road anomaly detection rely on vehicle dynamics, vibration analysis, or multi-sensor fusion 
(e.g., accelerometers, gyroscopes, and GPS)~\cite{salman2023deep,yin2024road,aguilar2025road}. 
While these approaches can effectively capture the vehicle’s response to road irregularities, 
they are inherently reactive, as they measure excitations only after the vehicle has encountered the disturbance. 
Consequently, they lack the ability to anticipate upcoming anomalies such as speed bumps and potholes, 
which fundamentally limits their applicability for predictive suspension control in real-time embedded systems.
To overcome the limitations of reactive suspension control, 
Fig.~\ref{fig:0} illustrates a vision-based preview strategy with adjustable damping 
for anticipating road irregularities such as speed bumps and potholes.

\input{Fig0.tex}

Recent advances in deep learning have enabled camera vision to emerge as an efficient and informative approach 
for proactive detection of road anomalies such as speed bumps and potholes. 
Compared to methods relying on acceleration signals, vision-based approaches provide earlier and more reliable detection, 
supplying critical preview information for real-time suspension control. 
Within this context, the YOLO (You Only Look Once) family has been influential in balancing detection accuracy with inference speed~\cite{youwai2024yolo9tr}. 
Bu{\v{c}}ko et al.~\cite{buvcko2022computer} evaluated YOLOv3 for pothole detection under diverse lighting and weather conditions, 
showing that while Sparse R-CNN offered slightly higher robustness in extreme scenarios, 
YOLOv3 provided stable performance with lower computational demands, making it suitable for embedded deployment. 
YOLOv4 further introduced CSPDarknet and PANet for improved feature aggregation. 
Park et al.~\cite{park2021application} demonstrated that YOLOv4-tiny could detect potholes in real time on resource-constrained devices, 
achieving high inference speed, while ASAD et al.~\cite{asad2022pothole} implemented a lightweight YOLOv4-tiny model on a Raspberry Pi, 
achieving approximately 90\% accuracy at 31.76 FPS for edge-based pothole detection. 
YOLOv5 added anchor-free heads and auto-learned anchors to enhance detection flexibility~\cite{jocher2020ultralytics}, 
and YOLOv8 incorporated decoupled detection heads and improved data augmentation~\cite{sohan2024review}. 
Khan et al.~\cite{khan2024pothole} further customized YOLOv8 for autonomous driving scenarios, optimizing it for road anomaly detection. 
POT-YOLO~\cite{bhavana2024pot} extended YOLOv8 to detect complex pavement defects with high accuracy. 
Building on YOLOv11~\cite{khanam2024yolov11}, 
which introduces the C3k2 residual block and C2PSA attention for enhanced multiscale features and scalable deployment, 
we optimize its architecture and inference for real-time, accurate road anomaly detection on embedded suspension systems.

Despite advances in vision-based road anomaly detection, real-world deployment under dynamic driving conditions remains challenging. 
Small, distant targets and environmental factors, such as motion artifacts and illumination variability, reduce detection robustness, 
while limited computational resources constrain input resolution and model complexity. 
To address these issues, this work proposes a systematic approach that jointly optimizes network architecture, training strategy, and model quantization 
for efficient and accurate detection of speed bumps and potholes. 
The main contributions are summarized as follows:

\begin{itemize}
\item[1)] Architecture and modules:  
A lightweight and efficient detection head (LEDH) is introduced to reduce the computational overhead of the P2 layer while preserving accuracy for small and distant targets. 
In combination with GhostConv modules and VoVGSCSPC blocks in the backbone and neck, the architecture improves feature extraction efficiency and robustness under diverse road and environmental conditions.  
\item[2)] Training strategy:  
A comprehensive training pipeline is employed to enhance performance in dynamic scenarios. 
It incorporates NWD loss for precise localization, BCKD knowledge distillation for effective feature transfer, 
and data augmentation with Albumentations to simulate motion blur, illumination variations, and adverse weather.  
\item[3)] Model quantization:  
The model is quantized to FP16 and optimized with TensorRT, achieving 139.5 FPS on NVIDIA Jetson AGX Xavier, 
thereby meeting real-time requirements for embedded road condition perception.  
\end{itemize}
The remainder of this paper is organized as follows:  
Sect.~\ref{method} presents the proposed model architecture and the improved loss functions.  
Sect.~\ref{exp} details the experimental setup and evaluation results.  
Sect.~\ref{sec:embedded Experiment} discusses the model’s inference performance on the embedded platform.  
Finally, conclusions are drawn in Sect.~\ref{sec:conclusion}.
%%%%%%%%%%%%%%%%%%%%%%%%%%
\section{Methods}
\label{method}
\subsection{SBP-YOLO Architecture}
\label{sec:sbp-yolo}
\input{fig1.tex}
This study focuses on the requirements of intelligent suspension systems, 
where accurate detection of road surface anomalies such as potholes and speed bumps is essential for predictive control. 
To address the constraints of limited computational resources and real-time inference, 
we adopt YOLOv11n as the baseline and introduce a series of structural improvements to form the proposed SBP YOLO architecture, 
as illustrated in Fig.~\ref{fig:1}.

The backbone network is enhanced by replacing conventional convolutional layers with GhostConv modules in deeper stages. 
This substitution significantly reduces computational cost while maintaining sufficient feature extraction capability. 
To improve the representation of features across different spatial resolutions, 
a new module named VoVGSCSPC is introduced. 
This design integrates densely connected patterns inspired by VoVNet with a GSConv-based bottleneck structure, 
and it is applied in both the backbone and the detection head to enhance semantic feature propagation.

To increase sensitivity to small and distant objects, 
an additional early-stage P2 detection layer is incorporated. 
This layer allows the model to capture fine spatial details by leveraging features from the shallow stages of the network. 
In contrast to the standard three-level detection structure (P3 to P5), 
SBP YOLO employs a four-level detection strategy that includes feature maps from P2 (160 $\times$ 160), 
P3 (80 $\times$ 80), P4 (40 $\times$ 40), and P5 (20 $\times$ 20). 
The detection head also integrates a lightweight and efficient design, 
referred to as LEDH, 
which further reduces computational overhead while preserving detection accuracy. 
All feature maps are annotated with their spatial dimensions, such as (320, 320, 16), 
where the three values represent the height, width, and number of channels, respectively. 
This notation provides a clear overview of the scale and structure of each layer.

Overall, 
SBP YOLO extends the baseline YOLOv11n by incorporating GhostConv modules, 
the VoVGSCSPC structure, a dedicated early detection branch, 
and the LEDH. 
These enhancements are designed to improve the model's ability to capture fine spatial details, 
enhance feature representation across multiple scales, 
and reduce computational complexity. 
Such architectural improvements aim to enable accurate and efficient detection of road surface anomalies, 
supporting deployment in intelligent vehicle control and predictive suspension systems.

\subsection{GhostConv Module}
\label{sec:ghostconv}
In practical road anomaly detection, identifying small and sparsely distributed targets—such as distant speed bumps and surface depressions—remains challenging due to their limited spatial coverage and low contrast with the surrounding environment. Effective detection of such anomalies requires a large receptive field while maintaining computational efficiency, which is particularly important for embedded platforms with constrained processing resources.

To address this issue, we incorporate the Ghost convolution (GhostConv) module from GhostNet~\cite{han2020ghostnet} as a lightweight substitute for standard convolution (SC). GhostConv follows a two-stage strategy: an initial small-kernel convolution extracts primary features, which are then enriched by a depthwise convolution using a larger kernel (e.g., \(5 \times 5\)) to generate supplementary "ghost" features. The depthwise convolution operates on half of the intermediate channels, allowing expansion of the receptive field with minimal computational overhead.

Integrating GhostConv into the backbone of SBP-YOLO enhances spatial feature representation while maintaining low complexity. This enables more accurate recognition of fine-grained road anomalies and supports efficient inference on low-power devices. As shown in Fig.~\ref{fig:2}, GhostConv significantly reduces the number of parameters and floating point operations (FLOPs) without sacrificing feature representation capability.

Let \( c_1 \) and \( c_2 \) represent the input and output channels, \( k \) the kernel size, and \( H \times W \) the spatial resolution. The parameter count and FLOPs for standard convolution are given by:
\begin{equation}
P_{\text{SC}} = c_1 c_2 k^2, \quad
F_{\text{SC}} = 2 c_1 c_2 H W k^2.
\end{equation}

For GhostConv, the initial convolution uses kernel size \( k_m \), and the depthwise convolution uses \( k_c \). The corresponding parameter count and FLOPs are:
\begin{align}
P_g &= c_1 \cdot \frac{c_2}{2} \cdot k_m^2 + \frac{c_2}{2} \cdot k_c^2, \\
F_g &= 2 c_1 \cdot \frac{c_2}{2} \cdot H W k_m^2 + 2 \cdot \frac{c_2}{2} \cdot H W k_c^2.
\end{align}

The cost ratios relative to standard convolution are:
\begin{equation}
\frac{P_g}{P_{\text{SC}}} = \frac{k_m^2 + \frac{k_c^2}{c_1}}{2k^2}, \quad
\frac{F_g}{F_{\text{SC}}} = \frac{k_m^2 + \frac{k_c^2}{c_1}}{k^2}.
\end{equation}

Under typical settings (\(c_1 \in [64, 256]\), \(k_m = 3\), \(k_c = 5\)), 
GhostConv reduces parameters and FLOPs to 48\%–55\% of SC, 
demonstrating its efficiency for embedded detection. 
Beyond computational savings, Fig.~\ref{fig:2}c shows that the augmented backbone 
enhances responses to distant small-scale anomalies, indicating an improved receptive field 
and richer feature representation.
\input{fig2.tex}
\subsection{VoVGSCSPC Module}
\label{sec:vovgscspc}
\input{fig3.tex}
Depthwise separable convolution (DSConv) reduces computational cost by decomposing standard convolution (SC) into a depthwise convolution (DWConv) for spatial filtering and a pointwise convolution (PWConv) for channel mixing. 
However, as spatial and channel operations are decoupled, DWConv suffers from limited inter-channel interaction, leading to feature isolation across groups. 
To address this, ShuffleNet~\cite{ma2018shufflenet} introduced channel shuffle to enable cross-channel information flow, 
while MobileNet~\cite{qin2024mobilenetv4} adopted extensive \(1 \times 1\) convolutions for dense channel fusion. 
Despite their effectiveness, these methods either suffer from reduced expressiveness or increased computational overhead, and often underperform SC in complex tasks.

GSConv~\cite{li2024slim} mitigates these drawbacks by combining the efficiency of GhostConv with a hardware-friendly channel shuffle mechanism. 
It enables effective intra- and inter-group feature communication, alleviating the isolation issue inherent in DWConv. 
By enhancing feature representation with minimal overhead, GSConv achieves accuracy comparable to or better than SC in lightweight models, 
making it well-suited for mobile and embedded visual applications.

As shown in Fig.~\ref{fig:3}, GSConv first applies a SC for downsampling, 
followed by a DWConv on the resulting features. 
The outputs are concatenated and shuffled to produce the final feature map:
\begin{equation}
\begin{aligned}
F_{\mathrm{GSC}} = \mathrm{Shuffle}\bigl(\mathrm{Cat}(\alpha(X_{C_1})_{C_2/2}, \delta(\alpha(X_{C_1})_{C_2/2}))\bigr)_{C_2}
\end{aligned}
\end{equation}
where $X_{C_1}$ is the input with $C_1$ channels, 
$\alpha$ denotes SC, 
$\delta$ represents DWConv, and $F_{\mathrm{GSC}}$ indicate the output.

VoVGSCSPC~\cite{li2025yolov8s} extends GSConv with a cross-stage architecture, 
splitting the input into two branches: one with a convolution followed by a GSBottleneck, 
and another with a DWConv residual branch. 
Their outputs are fused as
\begin{equation}
\begin{aligned}
GSB_{\mathrm{out}} = F_{\mathrm{GSC}}(F_{\mathrm{GSC}}(\alpha(X_{C_1})_{C_1/2})) + \alpha(X)_{C_1/2}
\end{aligned}
\end{equation}
\begin{equation}
\begin{aligned}
VoVGSCSP{C}_{\mathrm{out}} = \alpha\bigl(\mathrm{Concat}(GSB_{\mathrm{out}}, \alpha(X_{C_1}))\bigr)
\end{aligned}
\end{equation}
where $GSB_{\mathrm{out}}$ and $VoVGSCSP{C}_{\mathrm{out}}$ denote intermediate and final outputs, respectively.

The VoVGSCSPC module, 
employed instead of the C3k2 modules in the backbone and neck, 
enhances multi-scale feature fusion and enriches semantic representation. 
This improvement enables the model to better capture both global and local context in images of potholes and speed bumps, 
thereby strengthening its perceptual and representational capabilities.
% \input{fig3.tex}
%%%%%%%%%%%%%%%%%%%%%%%%%%%%%%%%%%%%%%%%%%%%%%%%%%%%%%%%%%%%%%%%%%%%%%%%%%%%%%%%%%%%%
\subsection{Lightweight and efficiency detection head (LEDH)}
\label{sec:ledh}
In suspension preview control, 
road anomalies such as speed bumps and potholes often appear as small, 
low-saliency objects at a distance, 
posing challenging small-object detection problems. 
Although YOLOv11 improves performance by introducing a shallow P2 detection layer, 
each detection head still employs two consecutive $3\times3$ convolutions before the classification and regression branches, 
leading to considerable computational overhead and hindering deployment on embedded platforms. 

To address this limitation, we propose the LEDH, 
a lightweight detection head that reduces redundancy while maintaining accuracy on small targets. 
As shown in Fig.~\ref{fig:4}, for each scale $i$, the input feature ${c}_i$ is first transformed by two stacked grouped convolutions (GroupConv) 
with group count $g = c_i / 16$:
\begin{equation}
\begin{aligned}
F_i &= \mathrm{Conv}_g\big( \mathrm{Conv}_g({c}_i) \big), \\
H_i &= \mathrm{Concat}\left[
      \mathrm{Conv}_{1\times1}^{4r}(F_i),\ 
      \mathrm{Conv}_{1\times1}^{n_c}(F_i)
      \right],
\end{aligned}
\end{equation}
where $F_i$ denotes the shared transformed feature map 
and $H_i$ the detection head output containing bounding box regression logits and class probabilities. 
Here, $n_c$ and $r$ are the number of classes and the Distribution Focal Loss (DFL) bin count, respectively. 

Unlike YOLOv11, which applies separate convolutional blocks to the classification and regression branches, 
LEDH introduces a shared transformation stage with GroupConv, 
so that the enhanced feature representation is jointly utilized by both branches. 
This design eliminates redundant computation, significantly reducing parameters and FLOPs, 
while preserving spatial detail and semantic abstraction across scales. 
Subsequent lightweight $1\times1$ convolutions produce the final predictions, 
yielding an efficient detection head well suited for real-time edge deployment.
%%%%%%%%%%%%%%%%%%%%%%%%%%%%%%%%%%%%%%%%%%%%%%%%%%%%%%%%%%%%%%%%%%%%%%%%%%%%%%%%%%%%%%%%%%%%%%%%%%%%
\input{fig4.tex}
\subsection{Hybrid Loss Training Strategy}
\label{sec:loss}
In road scene images, 
small structures such as speed bumps and potholes often appear with low pixel occupancy and blurred boundaries. 
This makes IoU-based losses highly sensitive to minor localization errors and insufficient for robust supervision. 
To mitigate this problem, 
we introduce the normalized Wasserstein distance (NWD)~\cite{wang2021normalized} as a distribution-aware metric, 
which models both predicted and ground truth boxes as 2D Gaussian distributions.
NWD provides smoother gradients under positional shifts and better tolerance to scale variation. 
By combining NWD with conventional IoU loss, 
the model gains improved localization precision, 
especially for small objects.
Given bounding boxes \( A = (c{x_a}, c{y_a}, w_a, h_a) \) and \( B = (c{x_b}, c{y_b}, w_b, h_b) \), 
the squared 2-Wasserstein distance between their corresponding Gaussian distributions \(\mathcal{N}_a\) and \(\mathcal{N}_b\) is computed. 
Since this distance does not directly serve as a similarity measure, it is normalized as:
\begin{equation}
\mathrm{NWD}(\mathcal{N}_a, \mathcal{N}_b) = \exp \left( - \frac{\sqrt{W_2^2(\mathcal{N}_a, \mathcal{N}_b)}}{C} \right),
\end{equation}
where \( C \) is a dataset-specific scaling constant (e.g., 0.5) determined via ablation studies.
The final bounding box regression loss is defined as:
\begin{equation}
\begin{cases}
L_{\mathrm{NWD}} = 1 - \mathrm{NWD}(\mathcal{N}_a, \mathcal{N}_b) \\
L_{\mathrm{obj}} = (1 - \alpha) \cdot L_{\mathrm{NWD}} + \alpha \cdot L_{\mathrm{iou}}
\end{cases}
\end{equation}
where \( L_{\mathrm{NWD}} \) represents the Wasserstein-based loss, 
\( L_{\mathrm{iou}} \) is the standard IoU loss, and \( \alpha = \mathrm{IoU}_{\mathrm{ratio}} \in [0, 1] \) controls the loss weighting.
\section{Experiments and Analysis}
\label{exp}
\subsection{Experimental settings}
\label{sec:exp-env}
The experiments were conducted on Ubuntu 24.04 with an Intel® Xeon E5-2680 CPU (2.4\,GHz) 
and an NVIDIA Tesla V100 GPU (16\,GB). 
The software environment comprised PyTorch 1.13.1 and CUDA 11.7. 
The model was trained using the Adam optimizer with an initial learning rate of 0.001, 
batch size of 16, and 300 epochs. 
Input images were resized to $640 \times 640$ to enhance convergence and computational efficiency. 
Early stopping was employed, terminating training if validation performance did not improve 
for 30 consecutive epochs.
\subsection{Dataset construction}
\label{sec:dataset}
Existing road surface datasets provide usable samples of potholes and speed bumps, 
but often lack comprehensive coverage of diverse driving conditions, 
such as varying lighting, occlusion, and road textures. 
To address this, we construct a targeted dataset by aggregating and refining relevant samples from multiple public sources, 
including benchmark datasets~\cite{nienaber2015detecting,varma2018real,buvcko2022computer,peralta2023speed} 
and community platforms such as Kaggle and Roboflow Universe. 
The resulting dataset captures representative road anomalies across complex real-world scenes, 
offering a more diverse and challenging benchmark for training and evaluating road perception models. 
Representative examples are shown in Fig.~\ref{fig:5}.

To further enhance generalization, 
data augmentation is applied to simulate challenging weather conditions 
such as rain and snow using efficient image transformation techniques~\cite{buslaev2020albumentations}. 
Additional augmentations include random flipping and photometric adjustments (e.g., brightness, contrast, saturation), 
as illustrated in Fig.~\ref{fig:6}. 
The final dataset consists of 7,500 images, 
divided into 5,250 for training, 1,125 for validation, and 1,125 for testing.
\input{fig5.tex}
\input{fig6.tex}
\subsection{Evaluation indicators}
\label{sec:dataset}
To quantitatively assess detection performance, 
we adopt mean average precision (\( \mathrm{mAP} \)), 
including mAP at 50\% IoU (\( \mathrm{mAP}_{50} \)) 
and the average over IoU thresholds from 50\% to 95\% (\( \mathrm{mAP}_{50\text{-}95} \)). 
Model efficiency is evaluated using the number of parameters (Params), 
GFLOPs, frames per second (FPS), and inference latency. 
Higher mAP reflects better detection accuracy, 
while higher FPS and lower latency indicate stronger real-time capability. 
GFLOPs and Params quantify computational and memory costs, 
which are critical for deployment on edge devices.

The evaluation metrics are defined as:
\begin{equation}
P = \frac{TP}{TP + FP}, \quad
R = \frac{TP}{TP + FN}
\end{equation}
\begin{equation}
\mathrm{AP} = \int_0^1 P(R) \, dR, \quad
\mathrm{mAP} = \frac{1}{M} \sum_{i=1}^{M} \mathrm{AP}_i
\end{equation}
\noindent
where \( TP \), \( FP \), and \( FN \) denote true positives, false positives, and false negatives, respectively. 
\( P \) and \( R \) represent precision and recall, 
and \( M \) is the number of object categories.
%%%%%%%%%%%%%%%%%%%%%%%%%%%%%%%%%%%%%%%%%%%%%%%%%%%%%%%%%%%%%%%%%%%%%%%%%%%%%%%
\subsection{Ablation experiment}
\label{sec:ablation exp}
\input{table1.tex}%tab
Ablation studies in Table~\ref{tab:1} evaluate the effectiveness of each proposed module by incrementally integrating them into the YOLOv11n baseline. 
Each configuration includes all modules from the previous row plus one newly added component.

The baseline YOLOv11n model demonstrates fast inference (71.2 FPS) 
but relatively limited detection accuracy (\( \mathrm{mAP}_{50} = 81.2\% \), \( \mathrm{mAP}_{50\text{-}95} = 47.7\% \)), 
primarily due to the absence of low-level feature information critical for detecting small objects. 
Introducing a P2 detection head notably improves accuracy to \( \mathrm{mAP}_{50} = 83.9\% \), 
although this comes at the cost of increased computational complexity (FLOPs rise from 6.3G to 10.3G) and a drop in inference speed to 53.5 FPS.

To achieve a better accuracy-efficiency trade-off, 
we replace the standard P2 head with the proposed LEDH. 
This substitution yields further improvements in precision (+1.8\%) and recall (+1.1\%), 
while significantly reducing FLOPs by 35\% and improving speed to 69.6 FPS.

Subsequently, 
the incorporation of GhostConv reduces the FLOPs by an additional 7.5\% with minimal impact on detection accuracy. 
Building on this, 
the integration of the VoVGSCSPC backbone module not only reduces model complexity (to 5.8 GFLOPs and 2.04M Params), 
but also improves detection accuracy to \( \mathrm{mAP}_{50} = 84.6\% \).

Finally, the application of the NWD Loss further enhances the model's ability to localize small and irregular objects, 
achieving the highest precision (89.0\%) and 
overall detection performance (\( \mathrm{mAP}_{50} = 86.6\% \), \( \mathrm{mAP}_{50\text{-}95} = 51.1\% \)) among all configurations. 
These results validate that each proposed component contributes to performance gains, 
and the final model presents a compelling balance of speed, accuracy, and lightweight design, 
making it well-suited for real-time road anomaly detection in resource-constrained environments.
%%%%%%%%%%%%%%%%%%%%%%%%%%%%%%%%%%%%%%%%%%%%%%%%%%%%%%%%%%%%%%%%%%%%%%%%%%%%%%%%%%%%%
\subsection{Knowledge Distillation Experiment}
\label{sec:ablation model distill}
Knowledge Distillation (KD) transfers knowledge from a larger, 
well-trained teacher model to a lightweight student model, 
aiming to enhance generalization with minimal computational overhead.
The proposed KD framework adopts a hybrid strategy, 
integrating logit-level supervision using BCKD~\cite{wang2023bckd} and feature-level guidance via CWD~\cite{shu2021channel}.
Six corresponding intermediate layers are selected from both models to enable effective feature alignment. 
The detailed training hyperparameters are summarized in Table~\ref{tab:2}.
% \input{table2.tex}%tab
% \input{table3.tex}%tab
% \input{table4.tex}%tab
\input{table2.tex}
\input{table3.tex}%tab
In our study, 
a teacher model with similar architecture and S-level scale (20.6 GFLOPs, 7.42M Params) was employed to guide the training of the student model. 
As shown in Table~\ref{tab:3}, 
KD leads to consistent improvements across all metrics. 
In particular, \( \mathrm{mAP}_{50\text{-}95} \) increased by 3.5\% on the validation set and 3.4\% on the test set, 
indicating not only enhanced detection performance but also improved generalization. 
The comparable gains across both sets suggest that the student model benefits from the teacher’s guidance without exhibiting overfitting.
%
% \input{table4.tex}%tab
%%%%%%%%%%%%%%%%%%%%%%%%%%%%%%%%%%%%%%%%%%%%%%%%%%%%%
\subsection{Model comparison experiment}
\label{sec:model comparison}
To comprehensively evaluate detection quality and computational efficiency, 
we compare SBP-YOLO with representative YOLO variants (YOLOv5-YOLOv12) 
across standard performance and complexity metrics. 
As reported in Table~\ref{tab:4}, SBP-YOLO achieves an \( \mathrm{mAP}_{50} \) of 86.6\%, 
surpassing the YOLOv11n baseline by 5.4\%, while reducing model parameters, FLOPs, 
and storage by 20.9\%, 7.9\%, and 17.9\%, respectively.
With knowledge distillation, SBP-YOLO (Distilled) further improves detection accuracy, 
reaching 87.0\% \( \mathrm{mAP}_{50} \) and 54.6\% \( \mathrm{mAP}_{50\text{-}95} \), 
without introducing additional computational overhead. 
These results highlight the effectiveness of our architecture and training strategy 
in achieving a strong trade-off between accuracy and efficiency.
\input{table4.tex}%tab
To further illustrate model performance during training, 
Fig.~\ref{fig:7} presents the \( \mathrm{mAP}_{50} \) and bounding box loss curves over 300 epochs. 
SBP-YOLO demonstrates consistently higher detection accuracy and lower training loss compared to other YOLO variants, 
indicating superior convergence speed and training stability. 
These trends further validate the effectiveness and robustness of SBP-YOLO for real-world deployment.
\input{fig7.tex}
%%%%%%%%%%%%%%%%%%%%%%%%%%%%%%%%%%%%%%%%%%%%%%%%%%%%%%%%%%%%%%%%%
\input{fig8.tex}
\subsection{Performance Comparison}
\label{sec:model comparison}
As shown in Fig.~\ref{fig:8}a, 
SBP-YOLO exhibits improved detection capability for small-scale road features at extended distances, 
such as far-field speed bumps and potholes. 
These targets often appear with reduced resolution and weak texture cues, 
posing challenges to existing detection models. 
Compared to baseline YOLO variants, 
SBP-YOLO achieves higher accuracy and greater consistency in identifying such long-range and low-visibility objects.
Fig.~\ref{fig:8}b illustrates detection performance under varied illumination conditions, 
including sunset, nighttime, and intense sunlight. 
SBP-YOLO maintains stable accuracy across all scenarios. 
At night, it outperforms YOLOv5 by 9\%, 
and during sunset—where reduced contrast can hinder detection—SBP-YOLO shows more robust results than other models. 
Under strong sunlight that induces glare, 
SBP-YOLO successfully identifies distant potholes, whereas models like YOLOv8 occasionally miss the target.
To assess robustness under visually degraded environments, 
Fig.~\ref{fig:8}c reports results on augmented datasets simulating rain, snow, and motion blur. 
SBP-YOLO achieves 0.64 detection accuracy in simulated rain, 
outperforming YOLOv11n (0.47) and YOLOv5n (lowest). 
Similar improvements are observed under snow and motion blur conditions, 
demonstrating the model’s resilience across complex visual perturbations.
These results demonstrate that SBP-YOLO provides a reliable and efficient solution for real-time detection of road surface features such as potholes and speed bumps. 
By identifying these geometric irregularities in advance, 
the model supports early adjustment of suspension parameters through feedforward control, 
helping to improve ride comfort and vehicle stability. 
This capability is particularly beneficial on complex or uneven roads, 
where timely perception is essential for safe and adaptive suspension behavior.
%%%%%%%%%%%%%%%%%%%%%%%%%%%%%%%%%%%%%%%%%%%%%%%%%%%%%%%%%%%%%%%%%%%%%%%%%%%%%%%%%%%%%%%%%%%%%%%%%%%%

\section{Experiment in embedded device}
\label{sec:embedded Experiment}
\subsection{Experimental Environment}
\label{sec:embedded Environment}
All embedded inference experiments were conducted on the NVIDIA Jetson AGX Xavier platform. 
As illustrated in Fig.~\ref{fig:9}, 
the experimental setup comprises the AGX Xavier module along with its carrier board. 
Inference was executed via the command line, 
enabling real-time monitoring of processing speed through the system terminal. 
The proposed SBP-YOLO model was converted and deployed using TensorRT, 
with the aid of an open-source optimization tool for YOLO models~\cite{timmurphy}. 
The software stack was based on NVIDIA JetPack 5.1.5.
\input{fig9.tex}
%
\subsection{Inference Experiments}
\input{table5.tex}%tab
\input{fig10.tex}
To ensure a fair comparison, 
the baseline YOLOv11n model was augmented with a P2 detection head to align with the detection structure of SBP-YOLO.
% This modification enables both models to detect small-scale targets at early feature stages, 
% eliminating potential biases caused by architectural differences. 
As a result, the performance evaluation focuses solely on the effectiveness of the proposed improvements in SBP-YOLO, 
rather than differences in model capacity or detection granularity.
Table~\ref{tab:5} and Fig.~\ref{fig:10} report the inference performance across various precision modes, 
including INT8, FP16, and FP32. 
Under FP16, SBP-YOLO achieves a 15.39 FPS gain over the modified YOLOv11n; with FP32, 
the improvement is 13.52 FPS. 
Additionally, SBP-YOLO maintains a smaller model size across all configurations, reflecting its superior efficiency.
When comparing different quantization modes under identical inference conditions, 
INT8 yields lower accuracy and frequent missed detections. 
FP16 maintains accuracy close to FP32, 
while significantly reducing latency. 
Notably, SBP-YOLO under FP16 consistently detects both long-range speed bumps and complex potholes, 
demonstrating a lower missed detection rate and higher reliability in real-time deployment scenarios.

\section{Conclusion}
\label{sec:conclusion}
This paper presented SBP-YOLO, 
a lightweight framework for real-time detection of speed bumps and potholes to support predictive suspension control. 
Building on YOLOv11, 
it incorporates a P2 detection head with LEDH employing shared GroupConv transformations to enhance 
small-object sensitivity while reducing computational overhead, 
alongside GhostConv and VoVGSCSPC modules for efficient multi-scale feature extraction. 
A hybrid training strategy combining NWD loss, 
backbone-level knowledge distillation, 
and advanced data augmentation improves robustness and generalization, 
achieving a 5.8\% mAP gain over the YOLOv11n baseline. 
After TensorRT FP16 optimization, 
the model reaches 139.5 FPS on Jetson AGX Xavier, a 12.4\% speedup over P2-enhanced YOLOv11n 
while maintaining superior accuracy. 
These results demonstrate the framework’s effectiveness for real-time embedded deployment, 
enabling anticipatory perception and adaptive damping control, 
with future work exploring infrared integration and multimodal detection to further enhance robustness 
under diverse conditions.
%%%%%%%%%%%%%%%%%%%%%%%%%%%%%%%%%%%%%%%%%%

\bibliographystyle{unsrt}  

\bibliography{references}   % name your BibTeX data base

\vspace{1em} 
\noindent\textbf{Data Availability}\quad 
The dataset, source code, and trained weights are openly available at 
\url{https://github.com/chuanqi1997/SBP-YOLO}.
\end{document}

%% file: fig0.tex
\begin{figure*}[htbp]% For two-column wide figures use
\centering
\includegraphics[width=0.7\textwidth]{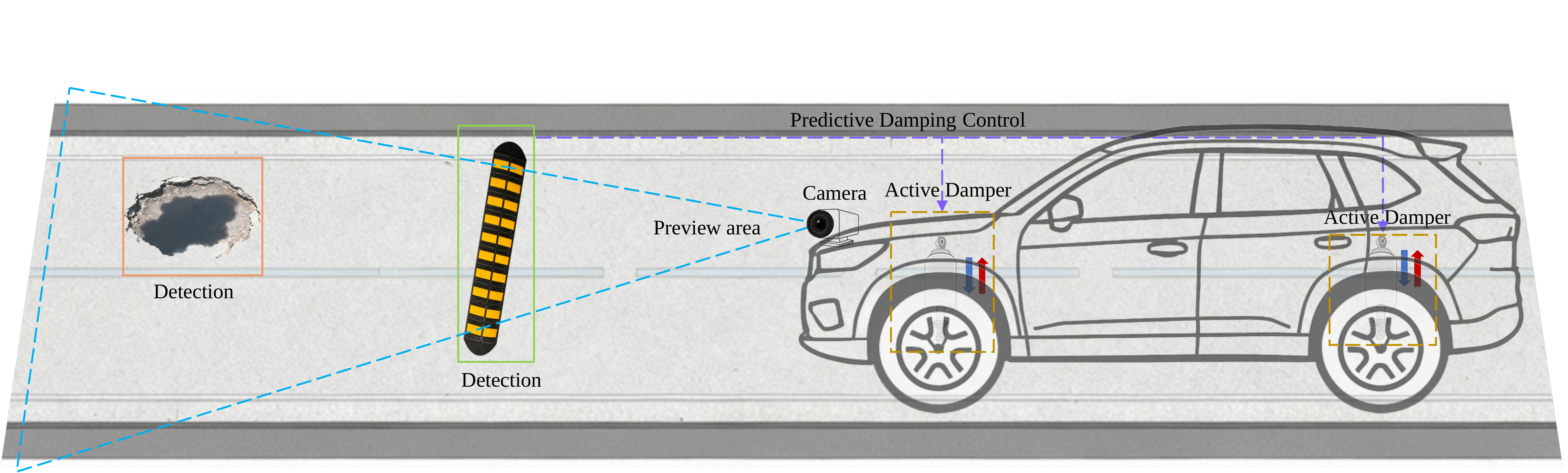}
\caption{Schematic of preview control principle with adjustable suspension for speed bumps and potholes}\label{fig:0}       % Give a unique label
\end{figure*}

%% file: fig1.tex
\begin{figure*}[htbp]% For two-column wide figures use
\centering
\includegraphics[width=0.5\textwidth]{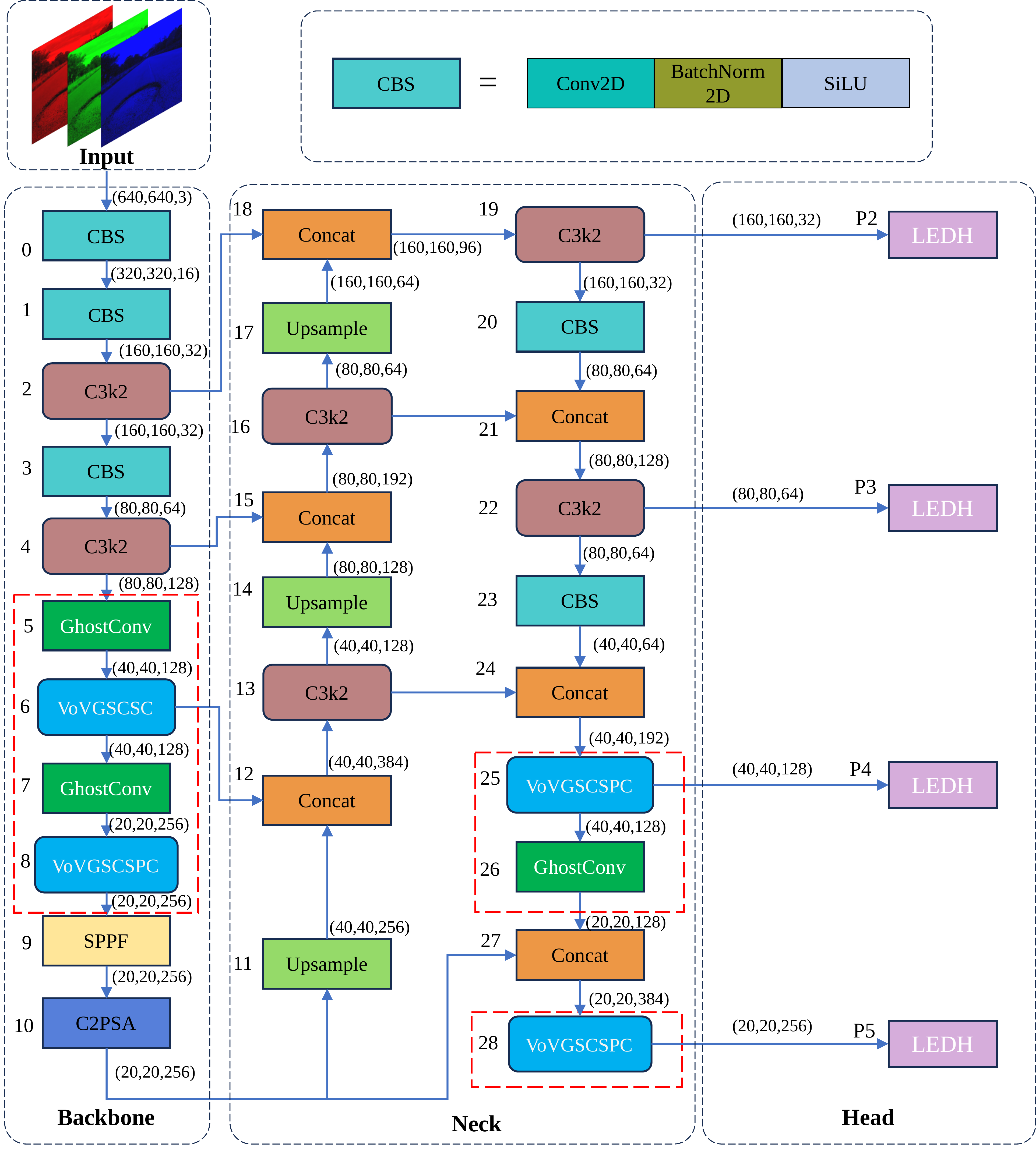}
% \caption{The framework of SBP-YOLO. SBP-YOLO consists of three parts: the Backbone, Neck, and Head. The neck receives four-scale (P2, P3, P4, and P5) inputs from the Backbone, performs multi-scale feature fusion, and connects with the head to conduct the final classification and regression.}
\caption{SBP-YOLO framework with Backbone, Neck, and Head. The Neck fuses multi-scale features (P2–P5) from the Backbone and passes them to the Head for classification and regression}
\label{fig:1}       % Give a unique label
\end{figure*}

%% file: fig2.tex
% \begin{figure}[htbp] %For one-column wide figures use
% \begin{figure*} %For one-column wide figures use
%   \centering
%   \includegraphics[width=0.7\textwidth]{Fig2.eps}  % 设置为页面宽度的80%
%   \caption{GhostConv structure diagram}
%   \label{fig:2}
% \end{figure*}
%%%%%%%%%%%%%%%%%%%%%%%%%%
% \begin{figure}[H] %For one-column wide figures use
\begin{figure}[htbp] %For one-column wide figures use
\centering
\includegraphics[width=0.5\textwidth]{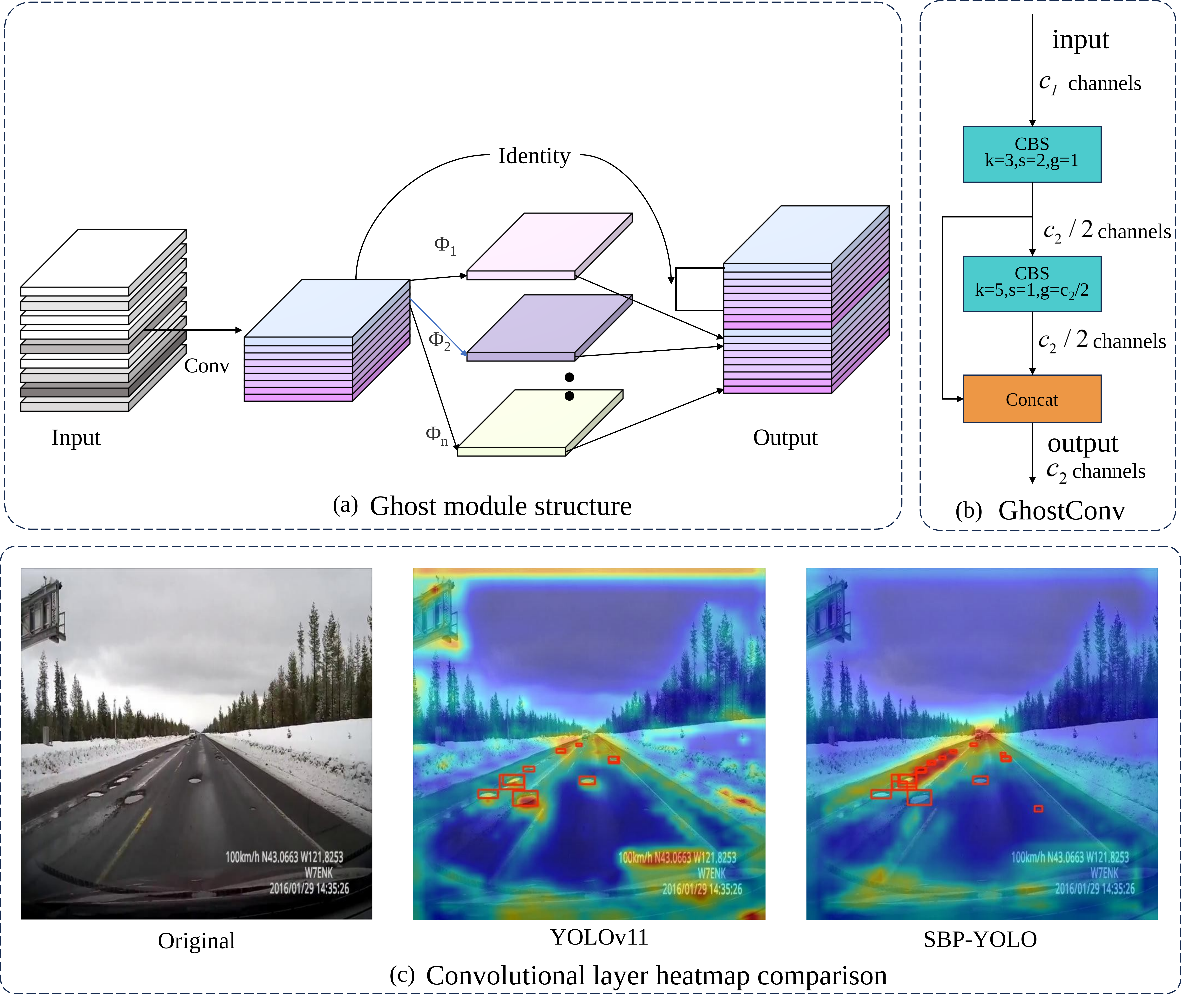}
% \includegraphics[width=0.48\textwidth]{Fig2_line.pdf}
% \caption{Ghost module and GhostConv structure diagram}
\caption{Illustration of the Ghost module design and its impact on feature activation. (a) Ghost module; (b) GhostConv architecture; (c) Comparison of backbone activation heat maps at the 5th and 7th convolutional layers.}
\label{fig:2}
\end{figure}

%% file: fig3.tex
% \begin{figure*}[!htbp] %For one-column wide figures use
%   \centering
%   \includegraphics[width=0.8\textwidth]{Fig3.eps}  % 设置为页面宽度的80%
%   \caption{VoVGSCSPCstructure diagram}
%   \label{fig:3}
% \end{figure*}
%%%%%%%%%%%%%%%%%%%%%%%%%%%%%%%%%%%%%
\begin{figure}[htbp] %For one-column wide figures use
  \centering
  \includegraphics[width=0.4\textwidth]{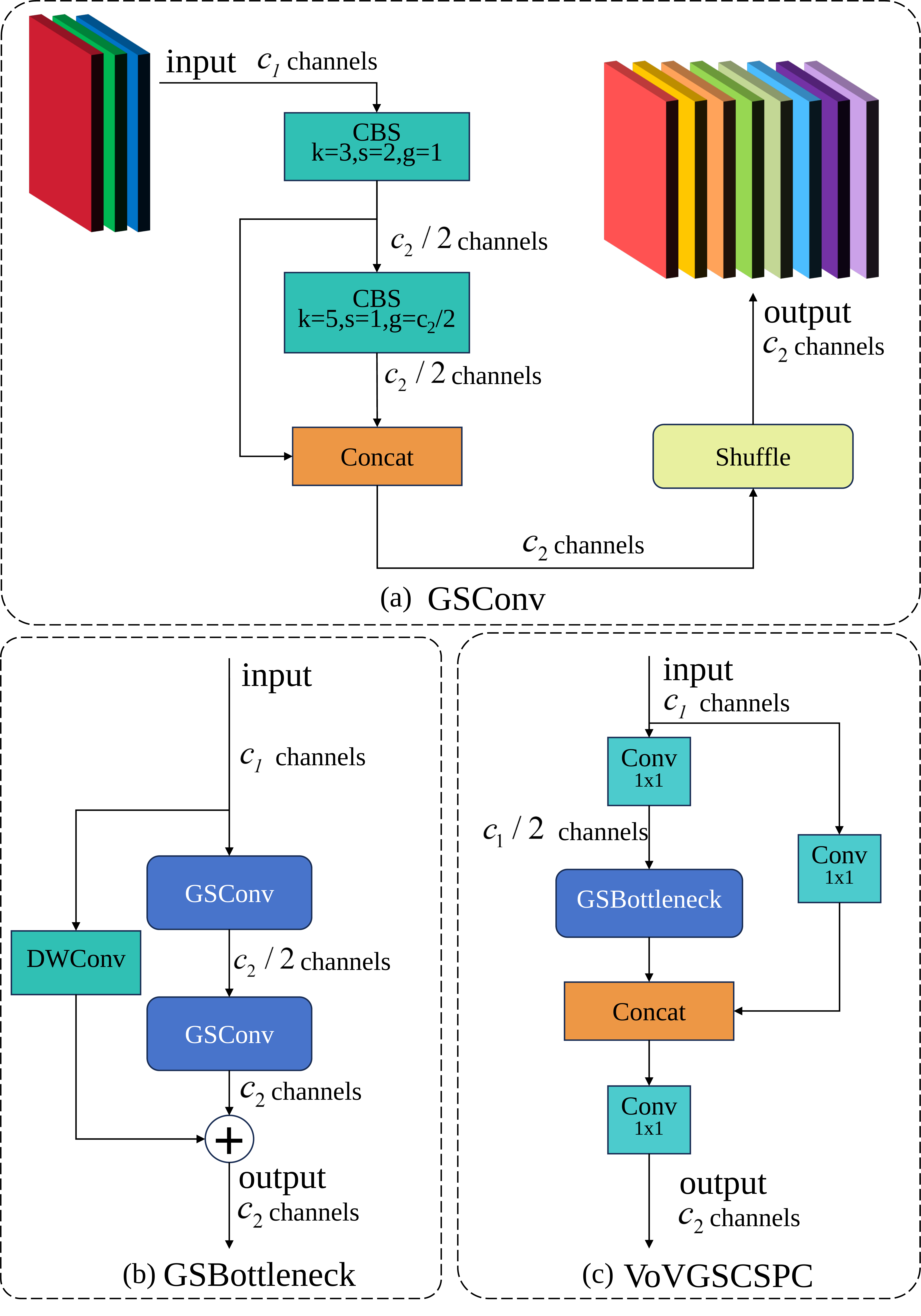}  % 设置为页面宽度的80%
  \caption{The flowchart of the VoVGSCSPC}
  \label{fig:3}
\end{figure}

%% file: fig4.tex
%%%%%%%%%%%%%%%%%%%%%%%%%%%%%%%%%%%%%%%%%%%%%%%%%%%%%%%%%%%%%%%%%%%%%%%%%%%%%%%%%%%%%%%%%%%%%%%%%%%%
\begin{figure*}[htbp] %For one-column wide figures use
\centering
\includegraphics[width=0.8\textwidth]{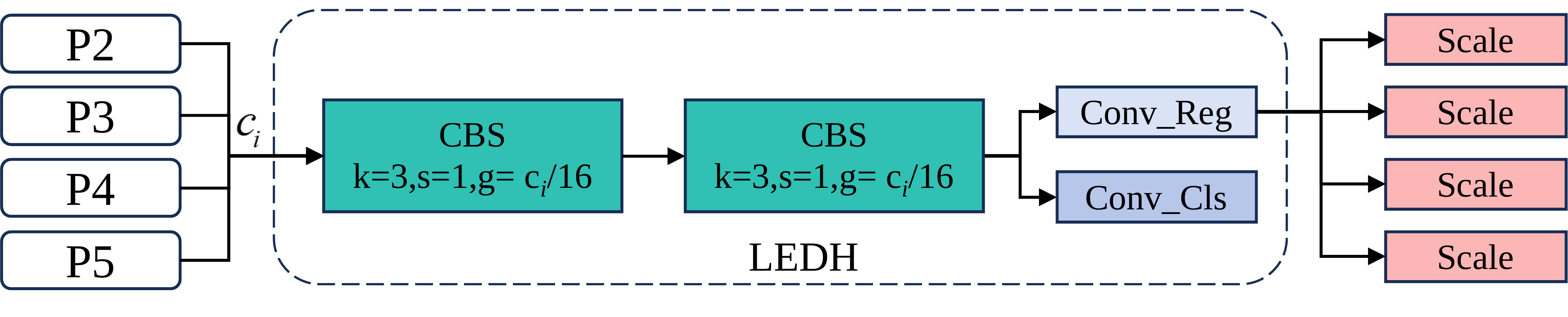}  % 设置为页面宽度的80%
\caption
{Structure of the proposed LEDH. 
Multi-scale features (P2--P5) are transformed by two stacked $3\times3$ GroupConv layers and then 
fed into parallel $1\times1$ convolution branches for classification and regression.
}
\label{fig:4}
\end{figure*}

%% file: fig5.tex
%%%%%%%%%%%%%%%%%%%%%%%%%%%%%%%%%%%%%%%%%%%%%%%%%%%%%%%%%%%%%%%%%%%%%%%%%%%%%%%%%%%%%%%%%%%%%%%%%%%%
\begin{figure*}[htbp] %For one-column wide figures use
  \centering
    \includegraphics[width=0.7\textwidth]{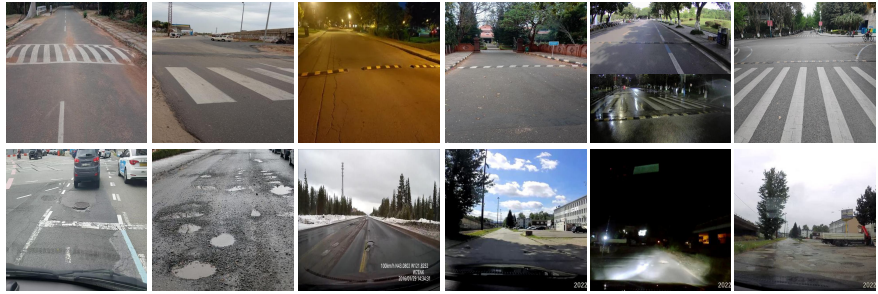}  % 设置为页面宽度的80%
  \caption{Images of potholes and speed bumps in the dataset}
  \label{fig:5}
\end{figure*}

%% file: fig6.tex
%%%%%%%%%%%%%%%%%%%%%%%%%%%%%%%%%%%%%%%%%%%%%%%%%%%%%%%%%%%%%%%%%%%%%%%%%%%%%%%%%%%%%%%%%%%%%%%%%%%%
\begin{figure*}[htbp] %For one-column wide figures use
  \centering
  \includegraphics[width=0.7\textwidth]{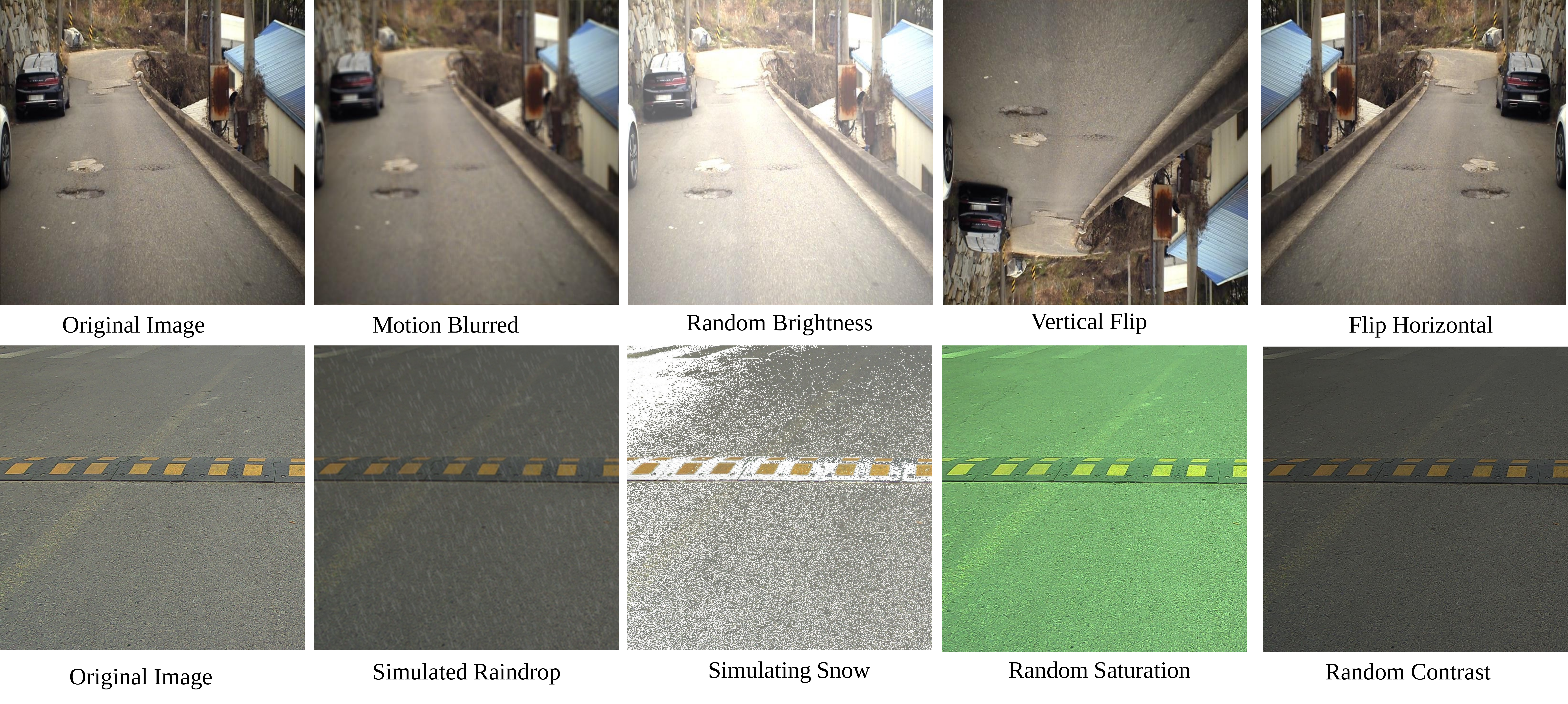}  % 设置为页面宽度的80%
  \caption{Comparison of the effects after some image data augmentation}
  \label{fig:6}
\end{figure*}

%% file: table1.tex
%%%%%%%%%%%%%%%%%%%%%%%%%%%%%%%%%%%%%%%%%%%%%%%%%%%%%%%%%%%%%%%%%%%%%%%%%%%%%%%%%%%%%%%%%%%%%%%%%%%%%%%%%%
% \begin{table*}[htbp]
% \caption{Ablation study of different module combinations}
% \label{tab:1}
% \centering
% \begin{tabular}{c|ccccc|ccccccc}
% \hline
% Model & A & B & C & D & E & Precision (\%) & Recall (\%) & \( \mathrm{mAP}_{50} \) (\%) & \( \mathrm{mAP}_{50\text{-}95} \) (\%) & GFLOPs & Params (M) & FPS \\
% \hline
% Baseline &  &  &  &  &  & 85.0 & 75.0 & 81.2 & 47.7 & 6.3 & 2.58 & \textbf{71.2} \\
% A        & \checkmark &  &  &  &  & 84.8 & 76.5 & 83.9 & 49.8 & 10.3 & 2.67 & 53.5 \\
% B        &  & \checkmark &  &  &  & 86.6 & 77.6 & 84.5 & 50.1 & 6.7 & 2.37 & 69.6 \\
% C        &  & \checkmark & \checkmark &  &  & 86.3 & 78.1 & 84.5 & 49.8 & 6.2 & 2.09 & 66.5 \\
% D        &  & \checkmark & \checkmark & \checkmark &  & 87.5 & 79.2 & 84.6 & 49.9 & 5.8 & 2.04 & 63.7 \\
% E        &  & \checkmark & \checkmark & \checkmark & \checkmark & \textbf{89.0} & \textbf{79.4} & \textbf{86.6} & \textbf{51.1} & \textbf{5.8} & \textbf{2.04} & 63.7 \\
% \hline
% \end{tabular}
% \end{table*}
%%%%%%%%%%%%%%%%%%%%%%%%%%%%%%%%%%%%%%%%%%%%%%%%%%%%%%%%%%%%%%%%%%%%%%%%%%%%%%%%%%%%%%
%消融实验
\begin{table}[htbp]
\caption{Ablation study by sequentially adding each module on top of the baseline. 
Each configuration includes all modules from the previous row plus the newly added component.}
\label{tab:1}
\centering
\begin{tabular}{l|cccccccc}
\hline
Model           & P (\%) & R (\%) & \( \mathrm{mAP}_{50} \) & \( \mathrm{mAP}_{50\text{-}95} \) & GFLOPs & Params (M) & FPS\textsubscript{b=1} \\
\hline
Baseline (YOLOv11n)            & 85.0 & 75.0 & 81.2 & 47.7 & 6.3  & 2.58 & \textbf{71.2} \\
+ P2 Head                      & 84.8 & 76.5 & 83.9 & 49.8 & 10.3 & 2.67 & 53.5 \\
+ LEDH (replacing P2)          & 86.6 & 77.6 & 84.5 & 50.1 & 6.7  & 2.37 & 69.6 \\
+ GhostConv                    & 86.3 & 78.1 & 84.5 & 49.8 & 6.2  & 2.09 & 66.5 \\
+ VoVGSCSPC                    & 87.5 & 79.2 & 84.6 & 49.9 & 5.8  & 2.04 & 63.7 \\
+ NWD Loss (Final)             & \textbf{89.0} & \textbf{79.4} & \textbf{86.6} & \textbf{51.1} & \textbf{5.8} & \textbf{2.04} & 63.7 \\
\hline
\end{tabular}
\end{table}

%% file: table2.tex
%%%%%%%%%%%%%%%%%%%%%%%%%%%%%%%%%%%%%%%%%%%%%%%%%%%%%%
% \begin{table}[htbp]
% \caption{Model distillation training hyperparameters}
% \label{tab:2}
% \centering
% \begin{tabular}{ll}
% \hline\noalign{\smallskip}
% \textbf{Hyperparameters} & \textbf{Configuration} \\
% \noalign{\smallskip}\hline\noalign{\smallskip}
% Epochs & 500 \\
% Batch size & 32 \\
% Optimizer & SGD \\
% Knowledge Distillation Loss Type & BCKD~\cite{wang2023bckd} \\
% Feature Loss Type & CWD~\cite{shu2021channel} \\
% Knowledge Distillation Loss Decay & Constant \\
% Logical Loss Ratio & 0.5 \\
% Feature Loss Ratio & 0.5 \\
% Teacher KD Layers & 12, 15, 18, 21, 24, 27 \\
% Student KD Layers & 12, 15, 18, 21, 24, 27 \\
% \noalign{\smallskip}\hline
% \end{tabular}
% \end{table}
\begin{table}[htbp]
\centering
\caption{Hyperparameter settings for the proposed Knowledge Distillation (KD) framework}
\label{tab:2}
\begin{tabular}{ll}
\hline\noalign{\smallskip}
\textbf{Hyperparameter} & \textbf{Value} \\
% \noalign{\smallskip}\hline\noalign{\smallskip}
Training Epochs & 500 \\
Batch Size & 32 \\
Optimizer & SGD \\
Logits Distillation Loss & BCKD \\
Feature Distillation Loss & CWD \\
Distillation Loss Schedule & Constant \\
Logits Loss Weight & 0.5 \\
Feature Loss Weight & 0.5 \\
Teacher Layers Used & 12, 15, 18, 21, 24, 27 \\
Student Layers Matched & 12, 15, 18, 21, 24, 27 \\
\noalign{\smallskip}\hline
\end{tabular}
\end{table}

%% file: table3.tex
\begin{table}[htbp]
\caption{Performance comparison of distilled model, teacher model, and non-distilled baseline on validation and test sets}
\label{tab:3}
\centering
\begin{tabular}{lcccc}
\hline\noalign{\smallskip}
\textbf{Validation Set} & P (\%) & R (\%) & \( \mathrm{mAP}_{50} \) & \( \mathrm{mAP}_{50\text{-}95} \) \\
\hline\noalign{\smallskip}
Baseline (Student)  & 89.0 & 79.4 & 86.6 & 51.1 \\
+ Distillation      & 90.2 & 80.7 & 87.0 & 54.6 \\
Teacher Model       & 91.0 & 81.4 & 87.4 & 53.8 \\
\hline
\textbf{Test Set} & P (\%) & R (\%) & \( \mathrm{mAP}_{50} \) & \( \mathrm{mAP}_{50\text{-}95} \) \\
\hline\noalign{\smallskip}
Baseline (Student)  & 88.6 & 78.2 & 86.3 & 51.5 \\
+ Distillation      & 89.9 & 81.8 & 87.7 & 54.9 \\
Teacher Model       & 89.9 & 82.0 & 87.8 & 54.1 \\
\hline
\end{tabular}
\end{table}

%% file: table4.tex
%%%%%%%%%%%%%%%%%%%%%%%%%%%%%%%%%%%%%%%%%%%%%%%%%%%%%%%%%%%%%%%%%%%%%%%%%%%%%%%%%%%%
% \begin{table*}[htbp]
% \caption{Model experiment comparison results}
% \label{tab:4}
% \centering
% \begin{tabular}{lccccccc}
% \hline\noalign{\smallskip}
% Model & P/\% & Recall & mAP@0.5 & mAP@0.5-0.95 & Params (M) & FLOPs (G) & Size (MB) \\
% \noalign{\smallskip}\hline\noalign{\smallskip}
% YOLOv12n~\cite{chileshe2025early} & 84.1 & 74.2 & 80.9 & 47.5 & 2.56 & 6.3 & 5.25 \\
% YOLOv11n & 85.0 & 75.0 & 81.2 & 47.7 & 2.58 & 6.3 & 5.21 \\
% YOLOv10n~\cite{wang2024yolov10} & 84.8 & 72.5 & 82.0 & 48.3 & 2.27 & 6.5 & 5.48 \\
% YOLOv9t~\cite{wang2024yolov9} & 84.7 & 75.3 & 81.5 & 48.3 & \textbf{1.76} & 6.6 & \textbf{3.95} \\
% YOLOv8n~\cite{sohan2024review} & 88.2 & 73.5 & 82.4 & 48.3 & 2.68 & 6.8 & 5.36 \\
% YOLOv6n~\cite{li2022yolov6} & 83.8 & 72.0 & 79.3 & 45.9 & 4.16 & 11.5 & 8.15 \\
% YOLOv5n~\cite{jocher2020ultralytics} & 83.4 & 74.5 & 80.6 & 46.1 & 2.18 & 5.8 & 4.43 \\
% SBP-YOLO & 89.0 & 79.4 & 86.6 & 51.1 & 2.04 & \textbf{5.8} & 4.28 \\
% \textbf{SBP-YOLO(distill)} & \textbf{90.2} & \textbf{80.7} & \textbf{87.0} & \textbf{54.6} & 2.04 & \textbf{5.8} & 4.28 \\
% \noalign{\smallskip}\hline
% \end{tabular}
% \end{table*}
\begin{table}[htbp]
\caption{Comparison of SBP-YOLO with state-of-the-art YOLO models in terms of accuracy and efficiency}
\label{tab:4}
\centering
\begin{tabular}{lccccccc}
\hline
Models & P (\%) & R (\%) & \( \mathrm{mAP}_{50} \) & \( \mathrm{mAP}_{50\text{-}95} \) & Params (M) & GFLOPs & Size (MB) \\
\hline
YOLOv5n~\cite{jocher2020ultralytics}     & 83.4 & 74.5 & 80.6 & 46.1 & 2.18 & 5.8 & 4.43 \\
YOLOv6n~\cite{li2022yolov6}              & 83.8 & 72.0 & 79.3 & 45.9 & 4.16 & 11.5 & 8.15 \\
YOLOv8n~\cite{sohan2024review}           & 88.2 & 73.5 & 82.4 & 48.3 & 2.68 & 6.8  & 5.36 \\
YOLOv9t~\cite{wang2024yolov9}            & 84.7 & 75.3 & 81.5 & 48.3 & \textbf{1.76} & 6.6  & \textbf{3.95} \\
YOLOv10n~\cite{wang2024yolov10}          & 84.8 & 72.5 & 82.0 & 48.3 & 2.27 & 6.5  & 5.48 \\
YOLOv11n~\cite{khanam2024yolov11}                                & 85.0 & 75.0 & 81.2 & 47.7 & 2.58 & 6.3  & 5.21 \\
YOLOv12n~\cite{chileshe2025early}        & 84.1 & 74.2 & 80.9 & 47.5 & 2.56 & 6.3  & 5.25 \\
SBP-YOLO (ours)                & 89.0 & 79.4 & 86.6 & 51.1 & 2.04 & \textbf{5.8}  & 4.28 \\
SBP-YOLO (distilled)           & \textbf{90.2} & \textbf{80.7} & \textbf{87.0} & \textbf{54.6} & 2.04 & \textbf{5.8} & 4.28 \\
\hline
\end{tabular}
\end{table}

%% file: fig7.tex
%%%%%%%%%%%%%%%%%%%%%%%%%%%%%%%%%%%%%%%%%%%%%%%%%%%%%%%%%%%%%%%%%%%%%%%%%%%%%%%%%%%%%%%%%%%%%%%%%%%
% \begin{figure*}[htbp] %For one-column wide figures use htbp
%   \centering
%   \includegraphics[width=0.7\textwidth]{Fig7.eps}  % 设置为页面宽度的80%
%   \caption{mAP0.5 and Box Loss comparison curves}
%   \label{fig:7}
% \end{figure*}
%%%%%%%%%%%%%%%%%%%%%%%%%%%%%%%%%%%%%%%%%%%%%%%%%%
\begin{figure}[htbp] %For one-column wide figures use htbp
  \centering
\includegraphics[width=0.5\textwidth]{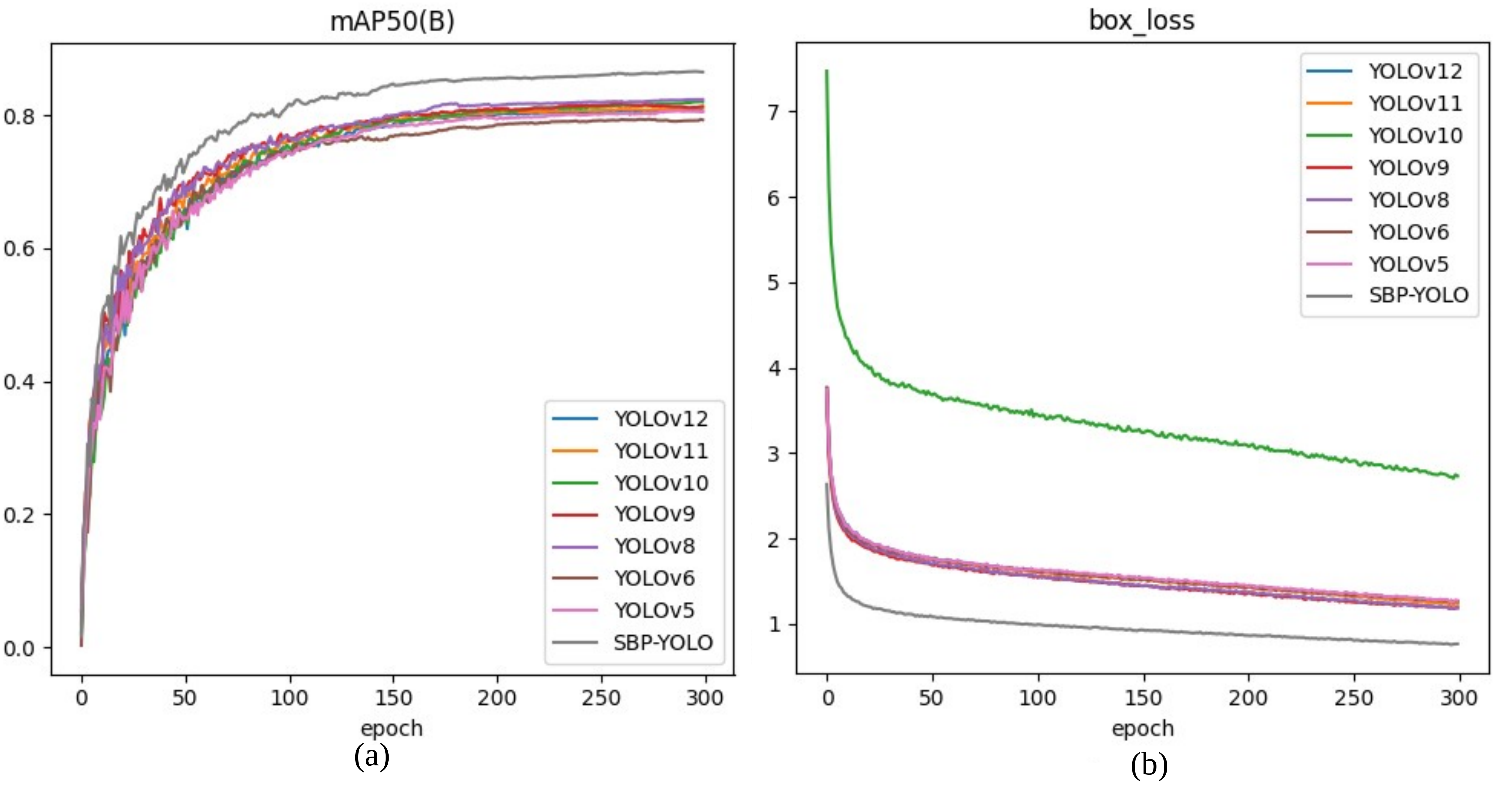}  % 设置为页面宽度的80%
\caption{Training curves over 300 epochs for different YOLO models: 
(a) \( \mathrm{mAP}_{50} \), 
(b) bounding box loss.}
  \label{fig:7}
\end{figure}

%% file: fig8.tex
%%%%%%%%%%%%%%%%%%%%%%%%%%%%%%%%%%%%%%%%%%%%%%%%%%%%%%%%%%%%%%%%%%%%%%%%%%%%%%%%%%%%%%%%%%%%%%%%%%%
% \begin{figure*}[htbp] %For one-column wide figures use
%   \centering
%   \includegraphics[width=0.8\textwidth]{Fig8.eps}  % 设置为页面宽度的80%
%   \caption{Qualitative comparison of SBP-YOLO and baseline models under various challenging conditions. 
%   (a) Detection of small-scale road features (e.g., distant speed bumps and potholes) at extended ranges; 
%   (b) Detection robustness under diverse illumination conditions, including sunset, nighttime, and strong sunlight; 
%   (c) Performance under adverse visual conditions, including simulated rain, snow, and motion blur.}
%   \label{fig:8}
% \end{figure*}
\begin{figure*}[!ht]
  \centering
  \includegraphics[width=0.75\textwidth]{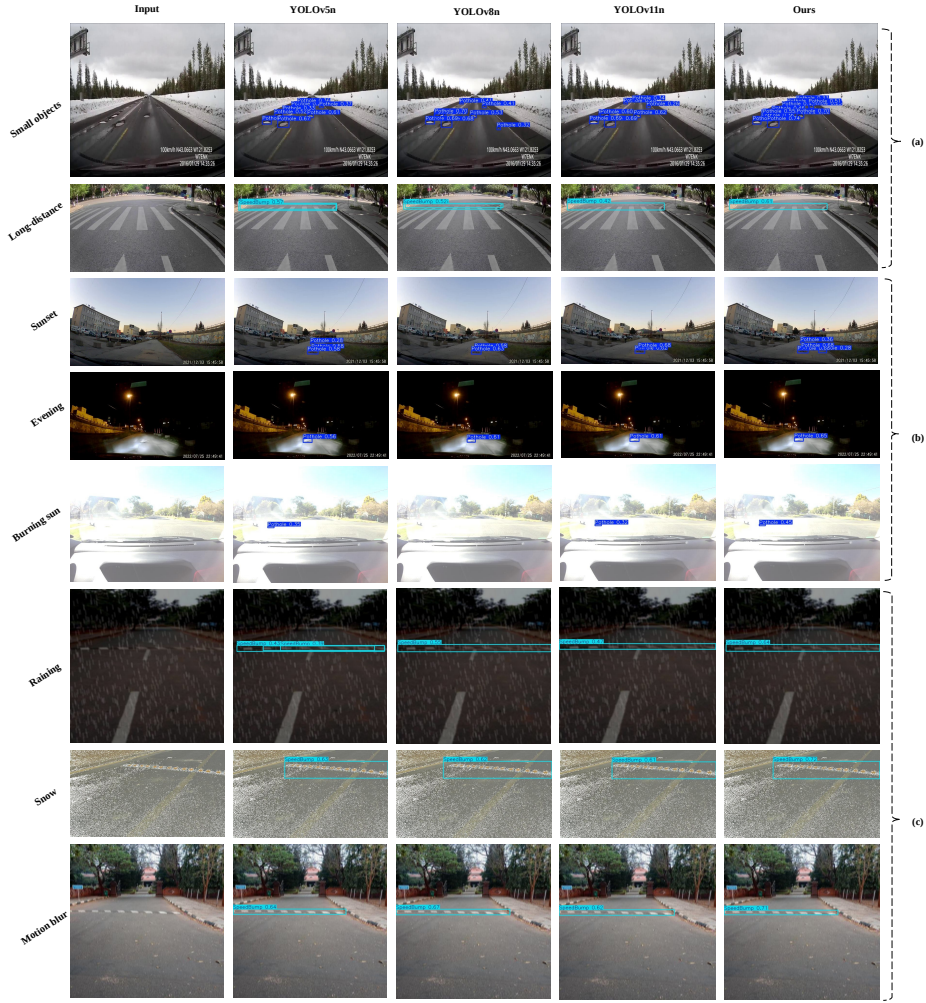}
  \caption{Qualitative comparison of SBP-YOLO and baseline models under challenging conditions: (a) detection of small distant road features; (b) robustness across varied lighting (sunset, night, strong sunlight); (c) performance under simulated rain, snow, and motion blur.}
  \label{fig:8}
\end{figure*}

%% file: fig9.tex
%%%%%%%%%%%%%%%%%%%%%%%%%%%%%%%%%%%%%%%%%%%%%%%%%%%%%%%%%%%%%%%%%%%%%%%%%%%%%%%%%%%%%%%%%%%%%%%%%%%
\begin{figure}[htbp] %For one-column wide figures use
  \centering
    \includegraphics[width=0.8\textwidth]{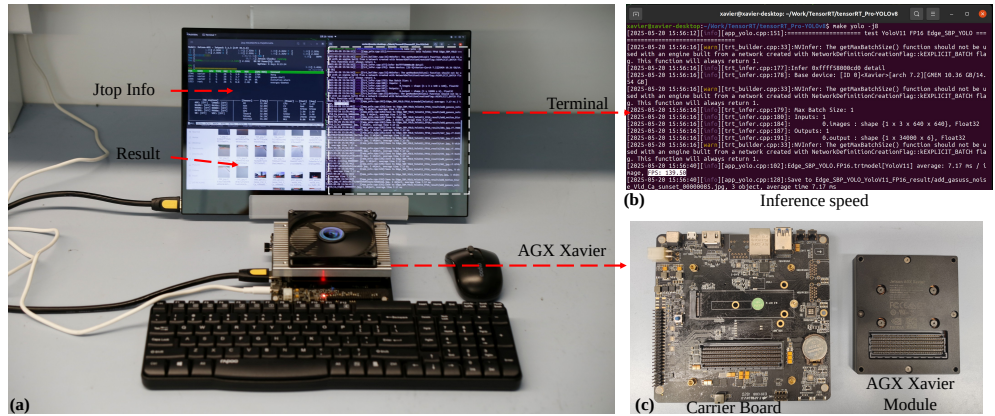}  % 设置为页面宽度的80%
  % \caption{Experimental Hardware Platform}
  % \caption{Experimental hardware setup: (a) experimental platform; (b) Jetson AGX Xavier-based hardware configuration; (c) display terminal for real-time inference results.}
  \caption{Experimental hardware setup: (a) experimental platform; (b) display terminal for real-time inference results; (c) Jetson AGX Xavier-based hardware configuration .}
  \label{fig:9}
\end{figure}
%%%%%%%%%%%%%%%%%%%%%%%%%%%%%%%%%%%%%%%%%%%%%%%%%
% \begin{figure*}[htbp] %For one-column wide figures use
%   \centering
%   \includegraphics[width=0.5\textwidth]{Fig9_two.eps}  % 设置为页面宽度的80%
%   \caption{Experimental Hardware Platform}
%   \label{fig:9}
% \end{figure*}

%% file: table5.tex
%%%%%%%%%%%%%%%%%%%%%%%%%%%%%%%%%%%%%%%%%%%%%%%%%%%%%%%%%%%%%%%%%%%%
\begin{table}[htbp]
\caption{Inference time, frame rate, and model size comparison between SBP-YOLO and YOLOv11n (with P2 head) under different quantization settings.}
\label{tab:5}
\centering
\begin{tabular}{lccc}
\hline\noalign{\smallskip}
Models & Infer Time (ms) & FPS & Weights \\
\noalign{\smallskip}\hline\noalign{\smallskip}
YOLOv11n (INT8) & 6.514 & 152.9 & 5.53 MB \\
YOLOv11n (FP16) & 8.06 & 124.11 & 7.42 MB \\
YOLOv11n (FP32) & 14.18 & 70.54 & 14.35 MB \\
SBP-YOLO (INT8) & \textbf{6.42} & \textbf{155.78} & \textbf{5.24 MB} \\
SBP-YOLO (FP16) & 7.17 & 139.50 & 6.18 MB \\
SBP-YOLO (FP32) & 11.9 & 84.06 & 9.63 MB \\
\noalign{\smallskip}\hline
\end{tabular}
\end{table}
% \begin{table}[htbp]
% \caption{Inference time, frame rate, and model size comparison between SBP-YOLO and YOLOv11n (with P2 head) under different quantization settings.}
% \label{tab:5}
% \centering
% \begin{tabular}{lccc}
% \hline\noalign{\smallskip}
% Model & Inference Time (ms) ↓ & FPS ↑ & Model Size ↓ \\
% \noalign{\smallskip}\hline\noalign{\smallskip}
% YOLOv11n (INT8) & 6.51 & 152.9 & 5.53 MB \\
% YOLOv11n (FP16) & 8.06 & 124.11 & 7.42 MB \\
% YOLOv11n (FP32) & 14.18 & 70.54 & 14.35 MB \\
% SBP-YOLO (INT8) & \textbf{6.42} & \textbf{155.78} & \textbf{5.24 MB} \\
% SBP-YOLO (FP16) & \textbf{7.17} & \textbf{139.50} & \textbf{6.18 MB} \\
% SBP-YOLO (FP32) & \textbf{11.90} & \textbf{84.06} & \textbf{9.63 MB} \\
% \noalign{\smallskip}\hline
% \end{tabular}
% \end{table}

%% file: fig10.tex
%%%%%%%%%%%%%%%%%%%%%%%%%%%%%%%%%%%%%%%%%%%%%%%%%%%%%%%%%%%%%%%%%%%%%%%%%%%%%%%%%%%%%%%%%%%%%%%%%%%
\begin{figure*}[htbp] %For one-column wide figures use
  \centering
  \includegraphics[width=0.8\textwidth]{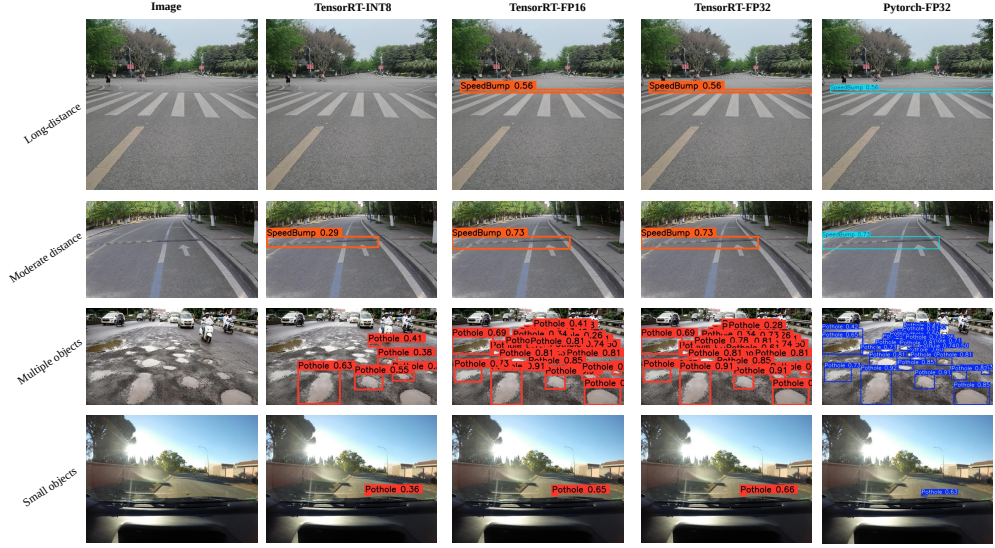}  % 设置为页面宽度的80%
  \caption{Comparison of recognition performance of different TensorRT quantization methods}
  \label{fig:10}
\end{figure*}

%% file: SBP-YOLO-arxiv.bbl
\begin{thebibliography}{10}

\bibitem{li2025robust}
Wei Li, Xiumei Du, Dongbin Xia, Jie Fu, and Miao Yu.
\newblock Robust h$\infty$ control for magnetorheological suspension system of all-terrain vehicles with parameter uncertainties and time delay.
\newblock {\em Vehicle System Dynamics}, pages 1--26, 2025.

\bibitem{yu2025research}
Miao Yu, Dongbin Xia, Wei Li, Gaowei Han, Xiumei Du, and Jie Fu.
\newblock Research on finite frequency robust h$\infty$ control of all-terrain vehicle magnetorheological suspension system.
\newblock {\em Mechanical Systems and Signal Processing}, 234:112803, 2025.

\bibitem{yu2024advances}
Min Yu, Simos~A Evangelou, and Daniele Dini.
\newblock Advances in active suspension systems for road vehicles.
\newblock {\em Engineering}, 33:160--177, 2024.

\bibitem{ferhath2024evolution}
Aadil~Arshad Ferhath and Kamalakkannan Kasi.
\newblock The evolution of damper technology for enhanced ride comfort and vehicle handling in vehicle suspension system.
\newblock {\em International Journal of Dynamics and Control}, 12(11):3908--3946, 2024.

\bibitem{achnib2023discrete}
Asma Achnib and Olivier Sename.
\newblock Discrete-time multi-model preview control: Application to a real semi-active automotive suspension system.
\newblock {\em Control Engineering Practice}, 137:105553, 2023.

\bibitem{wambold2009roughness}
James~C Wambold, Richard~A Zimmer, DL~Ivey, and Dean~L Sicking.
\newblock Roughness, holes, and bumps.
\newblock {\em Roadway Surface Discontinuities on Safety}, 11, 2009.

\bibitem{li2024explicit}
Wei Li, Huijun Liang, Dongbin Xia, Jie Fu, and Miao Yu.
\newblock Explicit model predictive control of magnetorheological suspension for all-terrain vehicles with road preview.
\newblock {\em Smart Materials and Structures}, 33(3):035037, 2024.

\bibitem{jung2025preview}
Jun~Young Jung and Chibum Lee.
\newblock Preview model predictive control of semi-active suspension for speed bump.
\newblock {\em International Journal of Automotive Technology}, 26(4):1115--1126, 2025.

\bibitem{wang2025road}
Guohong Wang, Farong Kou, Pengtao Liu, Wenhua Lv, and Longlong Xing.
\newblock Road preview method for active suspension based on reinforcement learning.
\newblock {\em Measurement Science and Technology}, 36(3):036206, 2025.

\bibitem{salman2023deep}
Amir Salman and Adnan~Noor Mian.
\newblock Deep learning based speed bumps detection and characterization using smartphone sensors.
\newblock {\em Pervasive and Mobile Computing}, 92:101805, 2023.

\bibitem{yin2024road}
Yunfei Yin, Wanli Fu, Xianyong Ma, Junpeng Yu, Xinkai Li, and Zejiao Dong.
\newblock Road surface pits and speed bumps recognition based on acceleration sensor.
\newblock {\em IEEE Sensors Journal}, 24(7):10669--10679, 2024.

\bibitem{aguilar2025road}
Abiel Aguilar-Gonz{\'a}lez and Alejandro Medina~Santiago.
\newblock Road event detection and classification algorithm using vibration and acceleration data.
\newblock {\em Algorithms}, 18(3):127, 2025.

\bibitem{youwai2024yolo9tr}
Sompote Youwai, Achitaphon Chaiyaphat, and Pawarotorn Chaipetch.
\newblock Yolo9tr: a lightweight model for pavement damage detection utilizing a generalized efficient layer aggregation network and attention mechanism.
\newblock {\em Journal of Real-Time Image Processing}, 21(5):163, 2024.

\bibitem{buvcko2022computer}
Boris Bu{\v{c}}ko, Eva Lieskovsk{\'a}, Katar{\'\i}na Z{\'a}bovsk{\'a}, and Michal Z{\'a}bovsk{\`y}.
\newblock Computer vision based pothole detection under challenging conditions.
\newblock {\em Sensors}, 22(22):8878, 2022.

\bibitem{park2021application}
Sung-Sik Park, Van-Than Tran, and Dong-Eun Lee.
\newblock Application of various yolo models for computer vision-based real-time pothole detection.
\newblock {\em Applied Sciences}, 11(23):11229, 2021.

\bibitem{asad2022pothole}
Muhammad~Haroon Asad, Saran Khaliq, Muhammad~Haroon Yousaf, Muhammad~Obaid Ullah, and Afaq Ahmad.
\newblock Pothole detection using deep learning: A real-time and ai-on-the-edge perspective.
\newblock {\em Advances in Civil Engineering}, 2022(1):9221211, 2022.

\bibitem{jocher2020ultralytics}
Glenn Jocher, Alex Stoken, Jirka Borovec, Liu Changyu, Adam Hogan, Laurentiu Diaconu, Jake Poznanski, Lijun Yu, Prashant Rai, Russ Ferriday, et~al.
\newblock ultralytics/yolov5: v3. 0.
\newblock {\em Zenodo}, 2020.

\bibitem{sohan2024review}
Mupparaju Sohan, Thotakura Sai~Ram, and Ch~Venkata Rami~Reddy.
\newblock A review on yolov8 and its advancements.
\newblock In {\em International Conference on Data Intelligence and Cognitive Informatics}, pages 529--545. Springer, 2024.

\bibitem{khan2024pothole}
Malhar Khan, Muhammad~Amir Raza, Ghulam Abbas, Salwa Othmen, Amr Yousef, and Touqeer~Ahmed Jumani.
\newblock Pothole detection for autonomous vehicles using deep learning: a robust and efficient solution.
\newblock {\em Frontiers in Built Environment}, 9:1323792, 2024.

\bibitem{bhavana2024pot}
N~Bhavana, Mallikarjun~M Kodabagi, B~Muthu Kumar, P~Ajay, N~Muthukumaran, and A~Ahilan.
\newblock Pot-yolo: Real-time road potholes detection using edge segmentation-based yolo v8 network.
\newblock {\em IEEE Sensors Journal}, 24(15):24802--24809, 2024.

\bibitem{khanam2024yolov11}
Rahima Khanam and Muhammad Hussain.
\newblock Yolov11: An overview of the key architectural enhancements.
\newblock {\em arXiv preprint arXiv:2410.17725}, 2024.

\bibitem{han2020ghostnet}
Kai Han, Yunhe Wang, Qi~Tian, Jianyuan Guo, Chunjing Xu, and Chang Xu.
\newblock Ghostnet: More features from cheap operations.
\newblock In {\em Proceedings of the IEEE/CVF conference on computer vision and pattern recognition}, pages 1580--1589, 2020.

\bibitem{ma2018shufflenet}
Ningning Ma, Xiangyu Zhang, Hai-Tao Zheng, and Jian Sun.
\newblock Shufflenet v2: Practical guidelines for efficient cnn architecture design.
\newblock In {\em Proceedings of the European conference on computer vision (ECCV)}, pages 116--131, 2018.

\bibitem{qin2024mobilenetv4}
Danfeng Qin, Chas Leichner, Manolis Delakis, Marco Fornoni, Shixin Luo, Fan Yang, Weijun Wang, Colby Banbury, Chengxi Ye, Berkin Akin, et~al.
\newblock Mobilenetv4: Universal models for the mobile ecosystem.
\newblock In {\em European Conference on Computer Vision}, pages 78--96. Springer, 2024.

\bibitem{li2024slim}
Hulin Li, Jun Li, Hanbing Wei, Zheng Liu, Zhenfei Zhan, and Qiliang Ren.
\newblock Slim-neck by gsconv: A lightweight-design for real-time detector architectures.
\newblock {\em Journal of Real-Time Image Processing}, 21(3):62, 2024.

\bibitem{li2025yolov8s}
Jun Li, Kaixuan Wu, Meiqi Zhang, Hengxu Chen, Hengyi Lin, Yuju Mai, and Linlin Shi.
\newblock Yolov8s-longan: a lightweight detection method for the longan fruit-picking uav.
\newblock {\em Frontiers in Plant Science}, 15:1518294, 2025.

\bibitem{wang2021normalized}
Jinwang Wang, Chang Xu, Wen Yang, and Lei Yu.
\newblock A normalized gaussian wasserstein distance for tiny object detection.
\newblock {\em arXiv preprint arXiv:2110.13389}, 2021.

\bibitem{nienaber2015detecting}
S~Nienaber, MJ~Booysen, and RS~Kroon.
\newblock Detecting potholes using simple image processing techniques and real-world footage.
\newblock Southern African Transport Conference, 2015.

\bibitem{varma2018real}
VSKP Varma, S~Adarsh, KI~Ramachandran, and Binoy~B Nair.
\newblock Real time detection of speed hump/bump and distance estimation with deep learning using gpu and zed stereo camera.
\newblock {\em Procedia computer science}, 143:988--997, 2018.

\bibitem{peralta2023speed}
Jos{\'e}-Eleazar Peralta-L{\'o}pez, Joel-Artemio Morales-Viscaya, David L{\'a}zaro-Mata, Marcos-Jes{\'u}s Villase{\~n}or-Aguilar, Juan Prado-Olivarez, Francisco-Javier P{\'e}rez-Pinal, Jos{\'e}-Alfredo Padilla-Medina, Juan-Jos{\'e} Mart{\'\i}nez-Nolasco, and Alejandro-Israel Barranco-Guti{\'e}rrez.
\newblock Speed bump and pothole detection using deep neural network with images captured through zed camera.
\newblock {\em Applied Sciences}, 13(14):8349, 2023.

\bibitem{buslaev2020albumentations}
Alexander Buslaev, Vladimir~I Iglovikov, Eugene Khvedchenya, Alex Parinov, Mikhail Druzhinin, and Alexandr~A Kalinin.
\newblock Albumentations: fast and flexible image augmentations.
\newblock {\em Information}, 11(2):125, 2020.

\bibitem{wang2023bckd}
Qi~Wang, Lu~Liu, Wenxin Yu, Shiyu Chen, Jun Gong, and Peng Chen.
\newblock Bckd: block-correlation knowledge distillation.
\newblock In {\em 2023 IEEE International Conference on Image Processing (ICIP)}, pages 3225--3229. IEEE, 2023.

\bibitem{shu2021channel}
Changyong Shu, Yifan Liu, Jianfei Gao, Zheng Yan, and Chunhua Shen.
\newblock Channel-wise knowledge distillation for dense prediction.
\newblock In {\em Proceedings of the IEEE/CVF international conference on computer vision}, pages 5311--5320, 2021.

\bibitem{li2022yolov6}
Chuyi Li, Lulu Li, Hongliang Jiang, Kaiheng Weng, Yifei Geng, Liang Li, Zaidan Ke, Qingyuan Li, Meng Cheng, Weiqiang Nie, et~al.
\newblock Yolov6: A single-stage object detection framework for industrial applications.
\newblock {\em arXiv preprint arXiv:2209.02976}, 2022.

\bibitem{wang2024yolov9}
Chien-Yao Wang, I-Hau Yeh, and Hong-Yuan Mark~Liao.
\newblock Yolov9: Learning what you want to learn using programmable gradient information.
\newblock In {\em European conference on computer vision}, pages 1--21. Springer, 2024.

\bibitem{wang2024yolov10}
Ao~Wang, Hui Chen, Lihao Liu, Kai Chen, Zijia Lin, Jungong Han, et~al.
\newblock Yolov10: Real-time end-to-end object detection.
\newblock {\em Advances in Neural Information Processing Systems}, 37:107984--108011, 2024.

\bibitem{chileshe2025early}
Martin Chileshe, Mayumbo Nyirenda, and John Kaoma.
\newblock Early detection of sexually transmitted infections using yolo 12: A deep learning approach.
\newblock {\em Open Journal of Applied Sciences}, 15(4):1126--1144, 2025.

\bibitem{timmurphy}
Melody Zhou.
\newblock tensorrt pro-yolov8.
\newblock \url{https://github.com/Melody-Zhou/tensorRT_Pro-YOLOv8}.
\newblock Accessed July 31, 2025.

\end{thebibliography}
